\documentclass{article}

 \usepackage[preprint]{neurips_2025}

\usepackage[utf8]{inputenc} 
\usepackage[T1]{fontenc}    
\usepackage{hyperref}       
\usepackage{url}            
\usepackage{booktabs}       
\usepackage{amsfonts}       
\usepackage{nicefrac}       
\usepackage{microtype}      
\usepackage{xcolor}         
\usepackage{amsmath}

\title{OpenEgo: A Large-Scale Multimodal Egocentric Dataset for Dexterous Manipulation}

\author{
  Ahad Jawaid\textsuperscript{1,2} \quad Yu Xiang\textsuperscript{1} \\
  \textsuperscript{1}Department of Computer Science, The University of Texas at Dallas \\
  \textsuperscript{2}Physical Automation, Inc. \\
  \texttt{ahad@physical.inc} \quad \texttt{yu.xiang@utdallas.edu}
}

\usepackage{graphicx}
\usepackage{pifont}
\newcommand{\cmark}{\textcolor{green!60!black}{\checkmark}}
\newcommand{\xmark}{\textcolor{red!70!black}{\ding{55}}}
\begin{document}

\maketitle

\begin{abstract}
    Egocentric human videos provide scalable demonstrations for imitation learning, but existing corpora often lack either fine-grained, temporally localized action descriptions or dexterous hand annotations. We introduce \textbf{OpenEgo}, a multimodal egocentric manipulation dataset with standardized hand-pose annotations and intention-aligned action primitives. OpenEgo totals \textbf{1107} hours across six public datasets, covering \textbf{290} manipulation tasks in \textbf{600+} environments. We unify hand-pose layouts and provide descriptive, timestamped action primitives. To validate its utility, we train language-conditioned imitation-learning policies to predict dexterous hand trajectories. OpenEgo is designed to lower the barrier to learning dexterous manipulation from egocentric video and to support reproducible research in vision--language--action learning. All resources and instructions will be released at \url{www.openegocentric.com}.
\end{abstract}

\section{Introduction}
\label{sec:intro}

Large-scale datasets have enabled generalization in language, speech, and vision \cite{diffusion,speechcodec}. In robotic manipulation, scaling remains difficult because collecting diverse, precisely annotated demonstrations is expensive \cite{openx,lfv-survey}. Prior work has explored learning from internet videos \cite{unipi}, but these sources differ from robot deployments in viewpoint, hand--object proximity, and motion patterns, creating a distribution gap that weakens transfer for dexterous control \cite{lfv-survey}. Egocentric video reduces this gap because hands and manipulated objects remain in frame from the actor’s perspective \cite{ego4d,epickitchen}.

Recent work highlights this direction. For instance, GR00T incorporates egocentric footage to mitigate visual domain gaps and improve action grounding for manipulation \cite{gr00t}. In parallel, imitation learning from egocentric demonstrations increasingly relies on 3D hand trajectories to supervise dexterous skills \cite{egomimic,humanpolicy,egodex,motiontracks}. Additionally, recent work on vision-language-action (VLA) models demonstrates a trend toward combining high-level action planning with low-level control policies, as seen in hierarchical approaches like $\pi_{0.5}$ \cite{pi0.5, hirobot}, where higher-level modules predict action primitives and lower-level modules execute them. Together, these trends reveal a gap: the lack of a single, large-scale egocentric dataset with unified hand joints and \emph{fine-grained, intention-aligned} language annotations.

We present \textbf{OpenEgo}, a consolidated dataset built from six public egocentric datasets spanning kitchen tasks, assembly, and daily activities (Table~\ref{tab:ego_datasets}) \cite{captaincook,hoi4d,holoassist,egodex,hot3d,hocap}. We convert hand poses into unified format and add intention-aligned action primitives with timestamps. We then train language-conditioned imitation-learning policies to predict future 3D hand trajectories.

\textbf{Contributions.} (1) A unified, large-scale egocentric manipulation dataset with standardized 21-joint hand poses and intention-aligned language annotations covering 290 manipulation tasks in 600+ environments. (2) An evaluation protocol focused on \emph{dexterous} 3D hand trajectory prediction. (3) To our knowledge, OpenEgo is the largest egocentric dataset with both dexterous hand annotations and fine-grained language action primitives suitable for training world models, hierarchical VLAs, and foundation VLM models, consolidating prior efforts into a unified format.

\section{Related Work}
\label{sec:related-works}

\textbf{Learning from video.}
Learning from video offers scale for imitation learning, but generic web sources suffer from distribution gap: they differ from robotic deployments in viewpoint, hand-object proximity, and motion, limiting transfer \cite{unipi,lfv-survey}. Egocentric video ensures hands and objects remain visible \cite{ego4d,epickitchen}. Beyond internet video, several methods learn directly from egocentric demonstrations with dexterous supervision. EgoDex aggregates large-scale hand-object interactions with 3D hand joints and frames trajectory prediction as a learning target \cite{egodex}. EgoMimic scales imitation learning from egocentric video using hand-centric supervision to produce fine-grained manipulation behavior \cite{egomimic}. Humanoid Policy leverages 3D hand pose trajectories to train dexterous controllers \cite{humanpolicy}. MotionTracks proposes structured trajectory representations for human-to-robot transfer suitable for few-shot imitation \cite{motiontracks}. Together, these works show that hand trajectory annotations and egocentric framing provide a practical route to robust manipulation learning.

\textbf{Vision-language-action models.}
The emergence of vision-language-action (VLA) models further emphasizes the value of egocentric footage. These models combine large-scale vision-language pretraining with action spaces relevant for manipulation \cite{openx,openvla}. A key example is GR00T, which incorporates egocentric footage to mitigate distribution gap and improve grounding of hand-object interactions \cite{gr00t}. Hierarchical VLA approaches show the utility of annotating data with fine-grained action primitives, where higher-level modules predict primitives and lower-level modules execute them \cite{hirobot,pi0.5}. This motivates datasets that unify temporally localized primitives with dexterous annotations.

\textbf{Egocentric datasets.}
Large egocentric corpora such as Ego4D \cite{ego4d} and EPIC-KITCHENS \cite{epickitchen} emphasize scale and diversity but generally lack the dexterous annotations required for manipulation. HoloAssist includes detailed interaction segments and 3D annotations but covers only 20 tasks \cite{holoassist}. EgoDex provides large-scale dexterous labels but no aligned language primitives \cite{egodex}. HOI4D captures 4D object-centric data with hand poses but at smaller scale \cite{hoi4d}. HOT3D and HO-Cap provide high-quality 3D annotations in lab setups but limited task diversity \cite{hot3d,hocap}. No single source combines scale, task diversity, fine-grained action primitives, and standardized dexterous annotations. OpenEgo fills this gap by consolidating six datasets into a unified format with intention-aligned language annotations and 21-joint hand trajectories.


\begin{figure}[t]
  \centering
  \includegraphics[width=\linewidth]{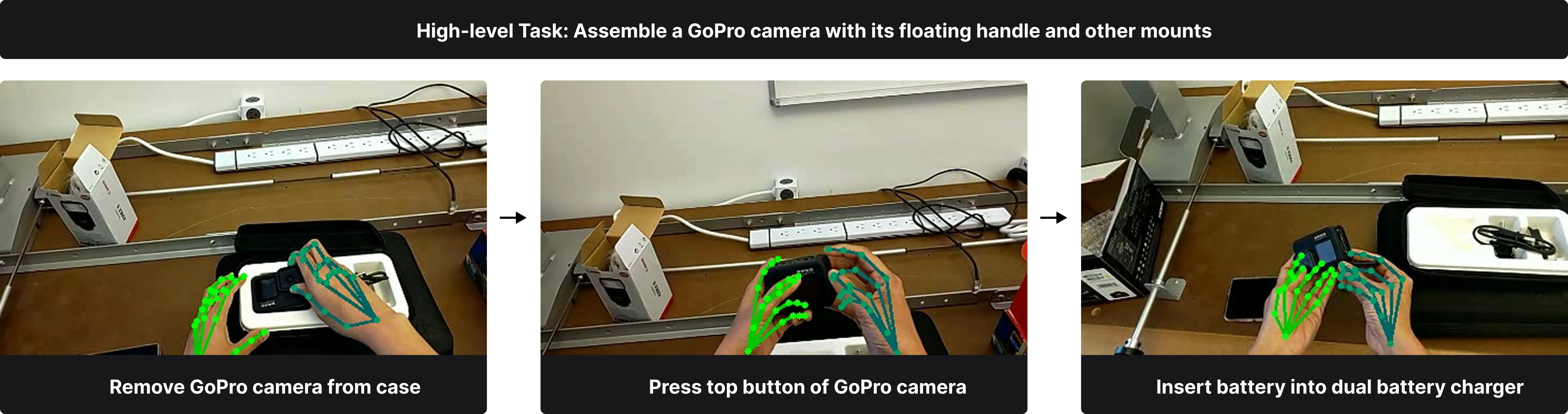}
  \caption{Illustration of high- and low-level task annotations in OpenEgo.}
  \label{fig:high_low_task}
  \vspace{-4mm}
\end{figure}

\section{OpenEgo Dataset}
\label{sec:openego}

{\setlength{\tabcolsep}{3.5pt}\renewcommand{\arraystretch}{1.03}}
\begin{table*}[t]
\centering
\footnotesize
\setlength{\tabcolsep}{4pt}
\renewcommand{\arraystretch}{1.05}
\begin{tabular}{@{}lccccccc@{}}
\toprule
\textbf{Dataset} & \textbf{Hours} & \textbf{\# Frames} & \textbf{\# Tasks} & \textbf{\# Record.} & \textbf{Fine-Grain} & \textbf{Dexterous} & \textbf{Coord. frame} \\
\midrule
CaptainCook4D \cite{captaincook} & 54   & 5.6M  & 24   & 200     & \xmark & \xmark & N/A \\
HOI4D \cite{hoi4d}               & 44   & 2.4M  & 16   & 4k      & \xmark & \cmark & Camera \\
HoloAssist \cite{holoassist}     & 166  & 17.9M & 20   & 2.2k    & \cmark & \cmark & World \\
EgoDex \cite{egodex}             & 829  & 90M   & 194  & 338k    & \xmark & \cmark & Camera \\
HOT3D \cite{hot3d}               & 13.3 & 3.7M  & 33   & 19      & \xmark & \cmark & World \\
HO-Cap \cite{hocap}              & 0.67 & 73k   & 3    & 64      & \xmark & \cmark & World \\
\midrule
OpenEgo (ours)                   & \textbf{1107} & \textbf{119.6M} & \textbf{290} & \textbf{344.5k} & \cmark & \cmark & Camera \\
\bottomrule
\end{tabular}
\caption{Egocentric datasets combined to form \textbf{OpenEgo}. “Fine-Grained Annotations” refers to temporal action segments; “Dexterous Annotations” refers to 3D multi-finger/hand-pose labels. The final column reports the source coordinate frame of hand-pose annotations. Distances (where applicable) are measured in meters.}
\label{tab:ego_datasets}
\end{table*}

\textbf{Overview.}
OpenEgo consolidates six egocentric sources into 1107 hours, 119.6M frames, 290 tasks, and 344.5k recordings (Table~\ref{tab:ego_datasets}). Environments span 10 kitchens and 610 indoor rooms, include 24 cooking recipes and 1.4k distinct objects, and involve at least 258 unique participants. The dataset contains both short- and long-form demonstrations.

\textbf{Hand joints.}
We align all hand poses to MANO’s 21-joint layout (4 joints per finger plus wrist) \cite{mano} and express final coordinates in the \emph{camera} frame. For sources lacking dexterous labels (CaptainCook4D), we detect 2D landmarks \cite{mediapipe} and back-project using dataset intrinsics and per-pixel depth from the RGB-D stream \cite{captaincook}. Given pixel $\mathbf{u}=(u,v,1)^\top$, depth $z$, and intrinsics $\mathbf{K}$, we recover camera-frame 3D by
\begin{equation}
\mathbf{X}_{c}\;=\; z\,\mathbf{K}^{-1}\,\tilde{\mathbf{u}},\quad \tilde{\mathbf{u}}=(u,v,1)^\top .
\end{equation}
When a dataset provides world-frame hand pose and calibrated extrinsics, we convert to the camera frame via the rigid transform supplied per frame:
\begin{equation}
\mathbf{X}_{c}\;=\;\mathbf{R}\,\mathbf{X}_{w}+\mathbf{t},\qquad (\mathbf{R},\mathbf{t}) \text{ from dataset extrinsics} \cite{holoassist,hot3d,hocap}.
\end{equation}
Per-dataset pipelines are as follows (all outputs are MANO-21 joints in camera frame):
\begin{itemize}
\item \textbf{CaptainCook4D} \cite{captaincook}: no native hand pose; detect 2D landmarks \cite{mediapipe} and back-project with depth to obtain 3D joints (camera frame).
\item \textbf{HOI4D} \cite{hoi4d}: uses released MANO parameters; we obtain MANO keypoints in the provided camera frame. For frames without poses, we fall back to landmark detection and depth back-projection as above \cite{hoi4d,mediapipe}.
\item \textbf{HoloAssist} \cite{holoassist}: hand poses are provided in the world frame; we convert to the camera frame using per-frame camera extrinsics supplied by the dataset \cite{holoassist}.
\item \textbf{EgoDex} \cite{egodex}: hand pose is in the camera frame; we map their 25-joint format to MANO-21 by dropping the four non-MANO joints and reindexing to MANO ordering \cite{egodex,mano}.
\item \textbf{HOT3D} \cite{hot3d}: MANO hand pose is defined in the world frame; we retrieve joints and transform to the camera frame using the dataset’s per-frame extrinsics \cite{hot3d}.
\item \textbf{HO-Cap} \cite{hocap}: MANO parameters are provided; we compute MANO keypoints and convert world-to-camera using the frame-wise extrinsics \cite{hocap}.
\end{itemize}
For all sources, we provide a binary visibility mask and retain missing joints when occluded or absent, following the dataset’s sensing and annotations \cite{captaincook,hoi4d,holoassist,egodex,hot3d,hocap}.

\textbf{Language primitives.}
We provide intention-aligned language annotations consisting of descriptive action primitives. Each primitive specifies the manipulated object and action, with absolute timestamps (\texttt{t\_start}, \texttt{t\_end}) from intention onset to completion. For example, primitives include \texttt{``navigate to desk''}, \texttt{``remove GoPro camera from case''}, or \texttt{``right hand unzips black camera case while left hand holds it''}. Manipulation entries include actor labels (\texttt{left\_hand}, \texttt{right\_hand}, \texttt{both\_hands}); navigation entries use \texttt{person} with a destination/object. Objects are named directly when identifiable, or described by attributes when uncertain. We annotate only directly observed actions and retain gaps between them. We also provide high-level task labels. This process is applied across all sources, including those with existing fine-grained annotations when descriptions differ or are incomplete \cite{holoassist}.


\section{Experiments}
\label{sec:experiment}

\textbf{Problem setup.} Each frame contains up to two hands with a unified 21-joint layout per hand (4 per finger plus wrist). We represent the state at time $t$ as stacked left/right hand joints $q_t \in \mathbb{R}^{42\times3}$ in the camera frame, with $(x,y,z)$ coordinates, and a binary visibility mask $m_t \in \{0,1\}^{42}$. Given an RGB observation $x_t$, a language prompt $\ell$ describing the intended manipulation, and the current joints $q_t$, the policy predicts the next $T$ actions $\hat q_{t+1:t+T}$, where $T$ is the prediction horizon used for training and evaluation.

\textbf{Policy and training.}
We adopt a ViLT policy following prior work \cite{atm,libero}. The model is trained with a masked mean-squared error objective,
\[
\mathcal{L} \;=\; \frac{1}{\sum_{\tau=1}^{T}\sum_{j} m_{t+\tau,j}} \sum_{\tau=1}^{T}\sum_{j} m_{t+\tau,j} \,\big\lVert \hat q_{t+\tau,j}- q_{t+\tau,j} \big\rVert_2^2,
\]
which ignores error on joints marked invisible by $m$. Due to computational limits, we train on a 0.1\% subset of OpenEgo and holding out 10\% of demonstrations for evaluation to avoid leakage. We optimize for 15{,}000 gradient steps using the AdamW \cite{adamw} optimizer and cosine annealing learning rate schedule with batch size 896 (\(\approx\)13.44M sampled training instances) on 2\,$\times$\,NVIDIA RTX 4090 GPUs. We follow the training framework of prior work \cite{atm,libero}. 

\textbf{Evaluation.}
We report sequence-prediction quality on the held-out split using only joints marked visible. We use three standard trajectory metrics: (i) \textbf{AED} — average Euclidean distance over the horizon; (ii) \textbf{FED} — Euclidean distance at the final step; and (iii) \textbf{DTW} — dynamic time-warping distance between predicted and reference joint sequences.

\textbf{Results.}
Table~\ref{tab:results} reports sequence prediction across horizons. Shorter horizons yield lower average and final errors; both AED and FED increase with horizon. DTW grows more steeply, reflecting sensitivity to timing misalignment. DTW grows faster than AED/FED because timing offsets accumulate over longer horizons, increasing alignment cost. The evaluation shows that models trained on OpenEgo can learn short-horizon dexterous motion effectively, with errors increasing smoothly as the prediction horizon extends. This indicates the dataset supplies a structured learning signal, where task difficulty scales smoothly with the prediction horizon.

\begin{table}[t]
\centering
\small
\renewcommand{\arraystretch}{1.25}
\begin{tabular}{lccc}
\toprule
\textbf{Horizon} & \textbf{Average Distance (AED) $\downarrow$} & \textbf{Final Distance (FED) $\downarrow$} & \textbf{DTW $\downarrow$} \\
\midrule
$8$ frames (0.5\,s @15\,fps) & 0.0491 & 0.0633 & 0.3918 \\
$16$ frames (1.0\,s) & 0.0714 & 0.0795 & 1.1284 \\
$32$ frames (2.0\,s) & 0.0893 & 0.0926 & 2.7150 \\
$64$ frames (4.0\,s) & 0.1045 & 0.1076 & 6.7975 \\
\bottomrule
\end{tabular}
\vspace{6pt}
\caption{Evaluation on OpenEgo for 3D hand-trajectory prediction at 15\,fps. AED: average Euclidean distance; FED: final-step Euclidean distance; DTW: dynamic time-warping distance.}
\label{tab:results}
\end{table}


\section{Conclusion}
\label{sec:conclusion}
 We introduced \textbf{OpenEgo}, a unified egocentric dataset for dexterous manipulation with standardized 21-joint hand poses in the camera frame and intention-aligned language primitives. We demonstrated that OpenEgo can support a language-conditioned imitation-learning policy for dexterous hand-trajectory prediction. To our knowledge, OpenEgo is the largest dataset consolidating dexterous and fine-grained language annotations for egocentric manipulation at scale.

\textbf{Limitations.} (1) Hand joints are missing in some frames due to occlusions or absent labels; visibility masks reflect this. (2) Language annotations are automatically generated and only partially verified; temporal drift can occur for long or ambiguous actions. (3) For sources lacking dexterous labels, 3D joint quality depends on the landmark estimator and depth sensing. (4) Experiments train on only 0.1\% of OpenEgo with a single architecture; results should not be interpreted as upper bounds.

\textbf{Ethics and Broader Impacts.} OpenEgo is built from publicly available datasets, each used under its respective license (Appendix~\ref{sec:licenses}). Potential risks include privacy concerns in egocentric video and potential misuse for surveillance. We mitigate these by adhering strictly to original dataset licenses, and providing clear terms of use at \url{www.openegocentric.com}.

\bibliographystyle{plain}
\bibliography{neurips_2025}


\appendix
\section{Licenses and Attribution}
\label{sec:licenses}

OpenEgo combines six publicly available egocentric datasets. We respect the license terms of each source and provide proper attribution in all releases:

\begin{itemize}
    \item \textbf{CaptainCook4D} \cite{captaincook}: Apache License 2.0.
    \item \textbf{HOI4D} \cite{hoi4d}: Creative Commons Attribution–NonCommercial 4.0 International (CC BY–NC 4.0).
    \item \textbf{HoloAssist} \cite{holoassist}: Community Data License Agreement (CDLA) v2.
    \item \textbf{EgoDex} \cite{egodex}: Creative Commons Attribution–NonCommercial–NoDerivatives 4.0 International (CC BY–NC–ND 4.0).
    \item \textbf{HOT3D} \cite{hot3d}: Creative Commons Attribution–ShareAlike 4.0 (CC BY–SA 4.0) / CC BY–NC–SA 4.0 (depending on subset).
    \item \textbf{HO-Cap} \cite{hocap}: Creative Commons Attribution 4.0 International (CC BY 4.0).
\end{itemize}

For datasets that permit redistribution, we will release processed annotations under the original license with attribution. Specifically for EgoDex, licensed under CC BY–NC–ND, our annotation files will be made available with permission from the authors, and users must first retrieve the underlying EgoDex data from the official source under its license terms.

All releases include license texts and attribution statements, ensuring compliance with each dataset’s terms of use.

\end{document}